\def\year{2020}\relax
\documentclass[letterpaper]{article} 
\usepackage{aaai20}  
\usepackage{times}  
\usepackage{helvet} 
\usepackage{courier}  
\usepackage[hyphens]{url}  
\usepackage{graphicx} 
\usepackage{graphicx}  
\usepackage[switch]{lineno}

\usepackage{booktabs}
\usepackage{amssymb}

\title{Fast Learning of Temporal Action Proposal via Dense Boundary Generator}

\author{Chuming Lin$^{*\dag}$ ~ ~ Jian Li$^{*\dag}$ ~ ~ Yabiao Wang$^{\dag}$  ~ ~ Ying Tai$^{\dag}$ \\
 \Large \textbf{~ ~ Donghao Luo$^{\dag}$ ~ ~ Zhipeng Cui$^{\dag}$ ~ ~ Chengjie Wang$^{\dag}$ ~ ~ Jilin Li$^{\dag}$ ~ ~ Feiyue Huang$^{\dag}$~ ~ Rongrong Ji$^{\ddag}$}\\
\textsuperscript{$^\dag$}Youtu Lab, Tencent ~~ 
\textsuperscript{$^\ddag$}Xiamen University, China\\
{\tt\small $^\dag$\{swordli, caseywang, yingtai, michaelluo, zhipengcui, jasoncjwang, jerolinli, garyhuang\}@tencent.com} \\ 
{\tt\small $^\dag$linchuming22@gmail.com, $^\ddag$rrji@xmu.edu.cn} \\
{\small \url{https://github.com/TencentYoutuResearch/ActionDetection-DBG}}
}

\begin{document}

\twocolumn[{%
\renewcommand\twocolumn[1][]{#1}%
\maketitle
}]


{
\renewcommand{\thefootnote}{\fnsymbol{footnote}}
\footnotetext[1]{indicates equal contributions.
This work was done when Chuming Lin was an intern at Tencent Youtu Lab.
}
}

\begin{abstract}
Generating temporal action proposals remains a very challenging problem, where the main issue lies in predicting precise temporal proposal boundaries and reliable action confidence in long and untrimmed real-world videos. In this paper, we propose an efficient and unified framework to generate temporal action proposals named Dense Boundary Generator (DBG), which draws inspiration from boundary-sensitive methods and implements boundary classification and action completeness regression for densely distributed proposals. In particular, the DBG consists of two modules: Temporal boundary classification (TBC) and Action-aware completeness regression (ACR). The TBC aims to provide two temporal boundary confidence maps by low-level two-stream features, while the ACR is designed to generate an action completeness score map by high-level action-aware features. Moreover, we introduce a dual stream BaseNet (DSB) to encode RGB and optical flow information, which helps to capture discriminative boundary and actionness features. Extensive experiments on popular benchmarks ActivityNet-1.3 and THUMOS14 demonstrate the superiority of DBG over the state-of-the-art proposal generator (e.g., MGG and BMN). 
\end{abstract}

\section{Introduction}
Generating temporal action proposals in video is a fundamental task, which serves as a crucial step for various tasks, like action detection and video analysis. In an optimal case, such proposals should well predict action intervals, with precise temporal boundaries and reliable confidence in untrimmed videos. Despite the extensive endeavors \cite{BSNeccv2018,BMNarxiv2019,MGGcvpr19}, temporal action proposal generation retains as an open problem, especially when facing action duration variability, activity complexity, blurred boundary, camera motion, background clutter and viewpoint changes in real-world scenarios.

\begin{figure}[t]
  \centering
  \includegraphics[trim={0 0 0 0mm},clip,width=1\linewidth]{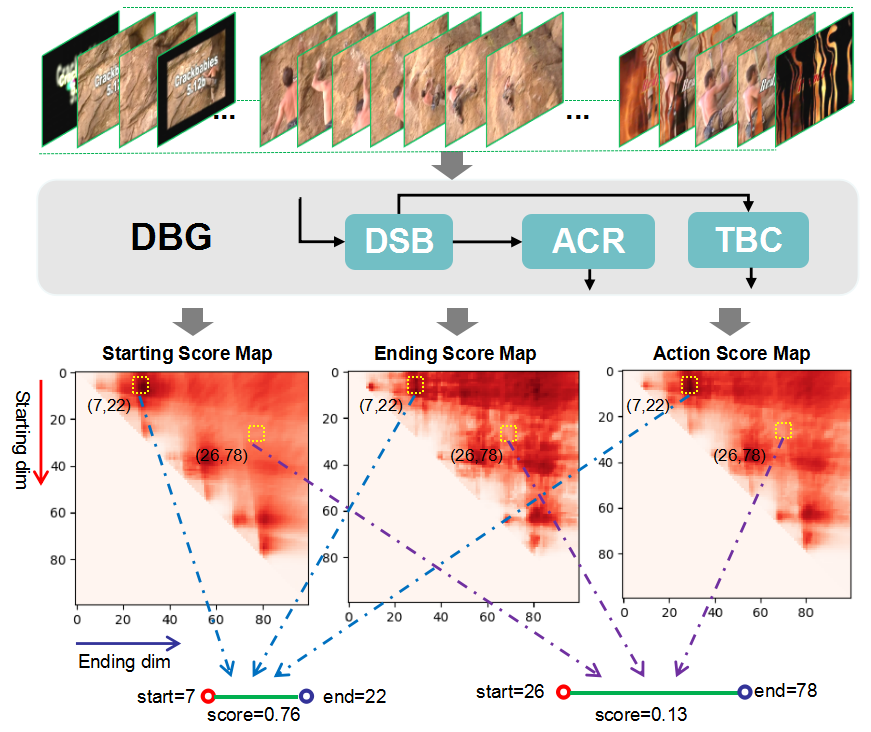}
  \caption{\small Overview of our proposed method. Given an untrimmed video, DBG densely evaluates all proposals by producing simultaneously three score maps: starting confidence score map, ending confidence score map and action completeness score map.
  }
\label{fig1}
\end{figure}

Previous works in temporal action proposals can be roughly divided into two categories: \emph{anchor based} ~\cite{SSTcvpr2017,Sparsecvpr2016,TURNiccv2017,S-CNNcvpr2016} and \emph{boundary based} ~\cite{TAGiccv2017,BSNeccv2018,BMNarxiv2019}. Anchor-based methods design a set of anchors at different scale for each video segment, which are regularly distributed over the video sequence. These candidate anchors are then evaluated by a binary classifier. However, anchor-based methods can not predict precise boundaries and are not flexible to cover multi-duration actions.

\begin{figure}[t]
  \centering
  \includegraphics[trim={0 0 0 0mm},clip,width=1\linewidth]{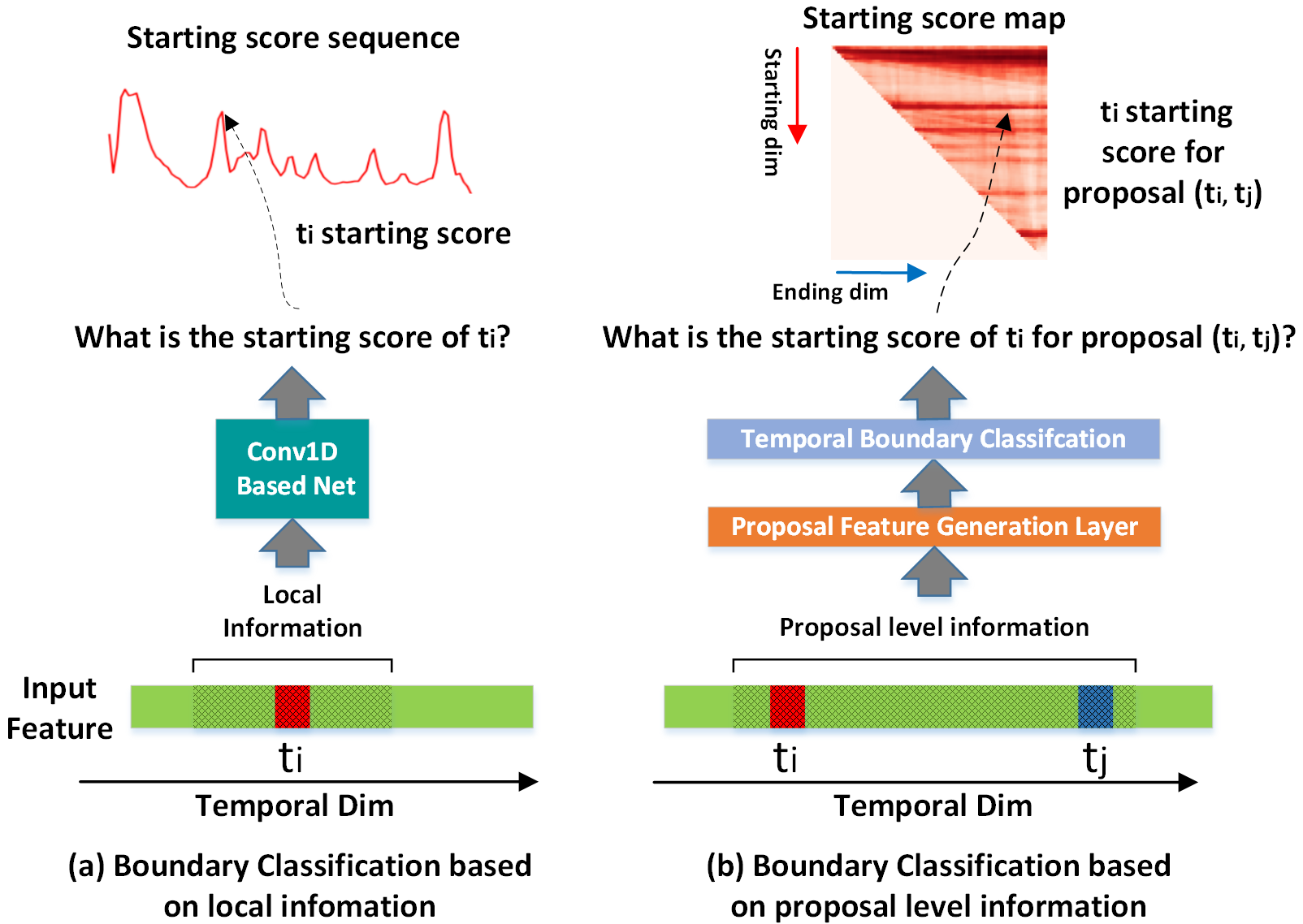}
  \caption{\small Boundary prediction comparison of (a) local information based and (b) global proposal information based methods.}
\label{fig2}
\end{figure}

Boundary-based methods evaluate each temporal location over the video sequence. Such local information helps to generate proposals with more precise boundaries and more flexible durations. As one of the pioneering works ~\cite{TAGiccv2017} groups continuous high-score regions as proposal by actionness scores. ~\cite{BSNeccv2018} adopts a two-stage strategy to locate locally temporal boundaries with high probabilities, and then evaluate global confidences of candidate proposals generated by these boundaries. To explore the rich context for evaluating all proposals, ~\cite{BMNarxiv2019} propose a boundary-matching mechanism for the confidence evaluation of proposals in an end-to-end pipeline. However, it drops actionness information and only adopts the boundary matching to capture low-level features, which can not handle complex activities and clutter background. Besides, different from our method shown in Fig. \ref{fig1}, it employs the same methods of ~\cite{BSNeccv2018} to generate boundary probability sequence instead of map, which lacks a global scope for action instances with blurred boundaries and variable temporal durations. Fig. \ref{fig2} illustrates the difference between local information and our global proposal information for boundary prediction.

To address the aforementioned drawbacks, we propose dense boundary generator (DBG) to employ global proposal features to predict the boundary map, and explore action-aware features for action completeness analysis. In our framework, a dual stream BaseNet (DSB) takes spatial and temporal video representation as input to exploit the rich local behaviors within the video sequence, which is supervised via actionness classification loss. DSB generates two types of features: Low-level dual stream feature and high-level actionness score feature. In addition, a proposal feature generation (PFG) layer is designed to transfer these two types of sequence features into a matrix-like feature. And an action-aware completeness regression (ACR) module is designed to input the actionness score feature to generate a reliable completeness score map. Finally, a temporal boundary classification (TBC) module is designed to produce temporal boundary score maps based on dual stream feature. These three score maps will be combined to generate proposals.

The main contributions of this paper are summarized as:
\begin{itemize}
\item We propose a fast and unified dense boundary generator (DBG) for temporal action proposal, which evaluates dense boundary confidence maps for all proposals.
\item We introduce auxiliary supervision via actionness classification to effectively facilitate action-aware feature for the action-aware completeness regression.
\item We design an efficient proposal feature generation layer to capture global proposal features for subsequent regression and classification modules.
\item Experiments conducted on popular benchmarks like ActivityNet-1.3 \cite{ANetcvpr2015} and THUMOS14 \cite{THUMOScviu2017} demonstrate the superiority of our network over the state-of-the-art methods.
\label{fig3}
\end{itemize}

\begin{figure*}[t]
\centering
\includegraphics[trim={0 0 0 0mm},clip,width=1\linewidth]{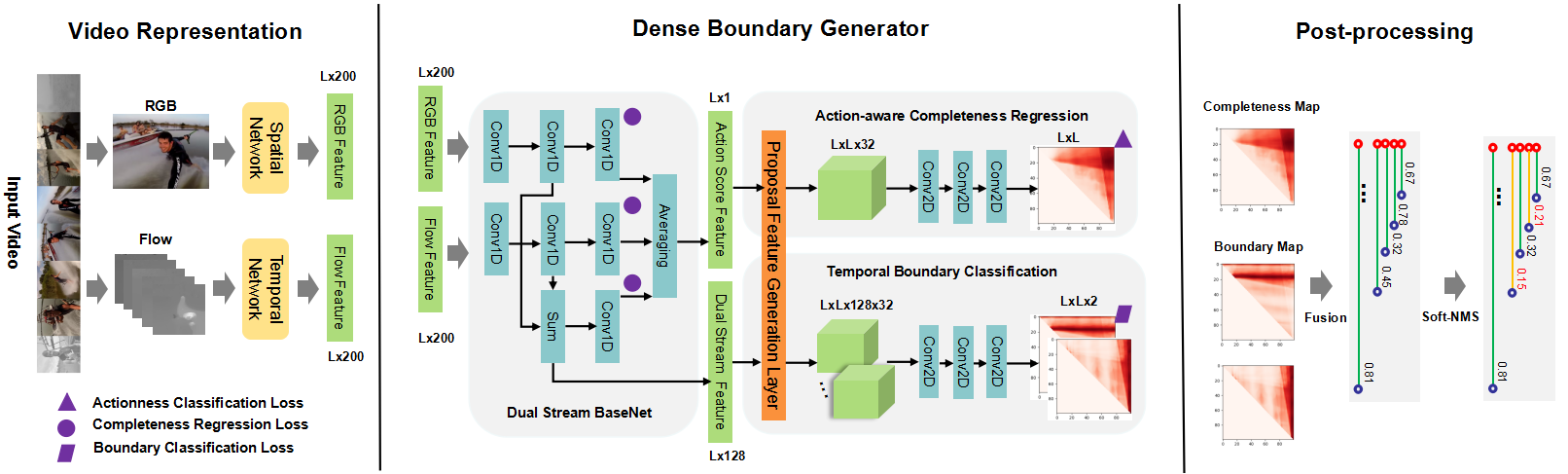}
\caption{ \emph{(a)Video Representation:} Spatial \& temporal network is used to encode video visual contents. \emph{(b)Dense Boundary Generator:} It contains Dual Stream BaseNet, Action-aware Completeness Regression branch and Temporal Boundary Classification branch. \emph{(c)Post-processing:} In this step, three score maps are fused and Soft-NMS is leveraged to generate proposals. }
\label{fig3}
\end{figure*}

\section{Related Work}

\textbf{Action recognition.} Early methods for video action recognition mainly relied on hand-crafted features such as HOF, HOG and MBH. Recent advances resort to deep convolutional networks to promote recognition accuracy. These networks can be divided into two patterns: \emph{Two-stream networks} ~\cite{CTNFcvpr2016,TCNnips2014,TGPcorr2015,TSNeccv2016}, and \emph{3D networks} ~\cite{LSF3Diccv2015,P3Diccv2017,I3Dcvpr2017}. Two-stream networks explore video appearance and motion clues by passing RGB image and stacked optical flow through ConvNet pretrained on ImageNet separately. Instead, 3D methods directly create hierarchical representations of spatio-temporal data with spatio-temporal filters.\\
\textbf{Temporal action proposal.}
Temporal action proposal aims to detect action instances with temporal boundaries and confidence in untrimmed videos. \emph{Anchor-based methods} generate proposals by designing a set of multi-scale anchors with regular temporal interval. The work in ~\cite{S-CNNcvpr2016} adopts C3D network ~\cite{LSF3Diccv2015} as the binary classifier for anchor evaluation. ~\cite{Sparsecvpr2016} proposes a sparse learning framework for scoring temporal anchors. ~\cite{TURNiccv2017} proposes to apply temporal regression to adjust the action boundaries. \emph{Boundary-based methods} evaluate each temporal location in video. ~\cite{TAGiccv2017} groups continuous high-score region to generate proposals by temporal watershed algorithm. ~\cite{BSNeccv2018} locates locally temporal boundaries with high probabilities and evaluate global confidences of candidate proposals generated by these boundaries. ~\cite{BMNarxiv2019} proposes a boundary-matching mechanism for confidence evaluation of densely distributed proposals in an end-to-end pipeline. MGG ~\cite{MGGcvpr19} combines anchor based method and boundary based method to accurately generate temporal action proposal.\\
\textbf{Temporal action detection.} The temporal action detection includes generating temporal proposal generation and recognizing actions, which can be divided into two patterns, \emph{i.e.}, one-stage ~\cite{SSADacmmm2017,GTANcvpr2019} and two-stage ~\cite{S-CNNcvpr2016,CBRbmvc2017,SSNiccv2017,RC3Diccv2017,TALcvpr2018}. The \emph{two-stage} method first generates candidate proposals, and then classifies these proposals. ~\cite{TALcvpr2018} improves two-stage temporal action detection by addressing both receptive field alignment and context feature extraction. For \emph{one-stage} method, ~\cite{SSADacmmm2017} skips the proposal generation via directly detecting action instances in untrimmed video. ~\cite{GTANcvpr2019} introduces Gaussian kernels to dynamically optimize temporal scale of each action proposal.

\section{Approach}
Suppose there are a set of untrimmed video frames $F=\{f_{t}\}^{l_{f}}_{t=1}$, where $f_{t}$ is the $t$-th RGB frame and $l_{f}$ is the number of frames in the video $V$. The annotation of $V$ can be denoted by a set of action instances $\psi_{g}=\{\varphi_{i}=(ts_{i},te_{i})\}^{N_{g}}_{i=1}$,  where $N_{g}$ is the number of ground truth action instances in video $V$, and $ts_{i}$, $te_{i}$ are starting and ending points of action instance $\varphi_{i}$. The generation of temporal action proposal aims to predict proposals $\psi_{p}=\{\varphi_{i}=(ts_{i},te_{i},p_i)\}^{N_{p}}_{i=1}$ to cover $\psi_{g}$ with high recall and overlap, where $p_i$ is the confidence of $\varphi_{i}$.

\subsection{Pipeline of our framework}
Fig. \ref{fig3} illustrates the proposed pipeline. In the phrase of video representation, spatial and temporal network are employed to encode video visual contents. The output scores of the two-stream network are used as RGB and flow features separately, which are fed into our dense boundary generator (DBG). DBG contains three modules: dual stream BaseNet (DSB), action-aware completeness regression (ACR) and temporal boundary classification (TBC). DSB can be regarded as a DBG backbone to exploit the rich local behaviors within the video sequence. DSB will generate two types of features: low-level dual stream feature and high-level actionness score feature. Actionness score feature is learned under auxiliary supervision of actionness classification loss, while dual stream feature is generated by late fusion of RGB and flow information. The proposal feature generation (PFG) layer transfers these two types of sequence features into a matrix-like feature. ACR will take actionness score features as input to produce an action completeness score map for dense proposals. TBC will produce temporal boundary confidence maps based on the dual stream features. ACR and TBC are trained by completeness regression loss and binary classification loss simultaneously. At last, the post-processing step generates dense proposals with boundaries and confidence by score map fusion and Soft-NMS.

\subsection{Video Representation}
To explore video appearance and motion information separately, we encode the raw video sequence to generate video representation by ~\cite{TSNeccv2016}, which contains spatial network for single RGB frame and temporal network for stacked optical flow field. We partition the untrimmed video frame sequence $F=\{f_{t}\}^{l_{f}}_{t=1}$ into snippets sequence $S=\{s_{t}\}^{l_{s}}_{t=1}$ by a regular frame interval $\delta$, where $l_{s}=l_{f}/\delta$. A snippet $s_{t}$ contains 1 RGB frame and 5 stacked optical flow field frames. We use output scores in the top layer of both spatial and temporal network to formulate the RGB feature $S_{t}$ and flow feature $T_{t}$. Thus, a video can be represented by a two-stream feature sequence $\{S_{t},T_{t}\}^{l_{s}}_{t=1}$. We set $l_{s}=L$ to keep the length of two-stream video feature sequence a constant.

\subsection{Dense Boundary Generator}
\textbf{Dual stream BaseNet.} The DBG backbone receives the spatial and temporal video feature sequences as input, and outputs actionness score feature and dual stream feature for ACR and TBC separately. DSB serves as the backbone of our framework, which adopts several one-dimensional temporal convolutional layers to explore local semantic information for capturing discriminative boundary and actionness features. As show in Tab. \ref{table1}, we use two stacked one-dimensional convolutional layers to exploit spatial and temporal video representation respectively, written by $sf=F_{conv12}(F_{conv11}(S))$, $tf=F_{conv22}(F_{conv21}(T))$. Then, following ~\cite{FeatureFusionicip2017}, we fuse $sf$, $tf$ by element-wise sum to construct low-level dual stream feature, denoted by $dsf=F_{sum}(sf,tf)$. Three convolutional layers will be adopted for $sf$, $tf$, $dsf$ separately to generate three actionness feature sequences $P^a=(F_{conv13}(sf), F_{conv23}(tf), F_{conv33}(dsf))$. In training, we use three auxiliary actionness binary classification loss to supervise $P^a$. In inference, three actionness feature sequence are averaged to generate high-level actionness score feature, which can be defined by $asf=F_{avg}(P^a)$.
\begin{table}[t]
\caption{The detail design of dual stream BaseNet (DSB), action-aware completeness regression (ACR) module and temporal boundary classification (TBC) module.}\smallskip
\centering
\resizebox{.95\columnwidth}{!}{
\smallskip\begin{tabular}{l|l|l|l|l|l}

\toprule
\multicolumn{6}{c}{ \textbf{DSB} } \\
\hline
layer  & kernel & output  & layer  & kernel & output\\
\hline
Conv1D$_{11}$  & 3 & L$\times$256 & Conv1D$_{21}$  & 3 & L$\times$ 256 \\
Conv1D$_{12}$  & 3 & L$\times$128 & Conv1D$_{22}$  & 3 & L$\times$ 128 \\
Sum  & \_ & L$\times$128 & Conv1D$_{33}$  & 1 & L$\times$1 \\
Conv1D$_{13}$  & 1 & L$\times$1 & Conv1D$_{23}$  & 1 & L$\times$ 1 \\
Averaging  & \_ & L$\times$1 & \-  & \- & \- \\
\toprule
\toprule
\multicolumn{3}{c|}{ \textbf{ACR} } & \multicolumn{3}{c}{ \textbf{TBC} } \\
\hline
layer  & kernel & output  & layer  & kernel & output\\
\hline
PFG  & \_ & L$\times$L$\times$32 & PFG  & \_ & L$\times$L$\times$32$\times$128\\
Conv2D$_{11}$ & 1$\times$1 & L$\times$L$\times$256 & Conv3D$_{21}$ & 1$\times$1$\times$32 & L$\times$L$\times$512\\
Conv2D$_{12}$ & 1$\times$1 & L$\times$L$\times$256 & Conv2D$_{22}$ & 1$\times$1 & L$\times$L$\times$256 \\
Conv2D$_{13}$ & 1$\times$1 & L$\times$L$\times$1 & Conv2D$_{23}$ & 1$\times$1 & L$\times$L$\times$2 \\
\toprule
\end{tabular}
}
\label{table1}
\end{table}\\
\textbf{Proposal feature generation layer.} The PFG layer is an efficient and differentiable layer that is able to generate temporal context feature for each proposal and make our framework be end-to-end trainable. For an arbitrary input feature $f^{in}$ whose shape is $L\times C$, the PFG layer is able to produce the proposal feature tensor whose shape is $L\times L\times N \times C$, which contains $L \times L$ proposal features $f^{p}$ whose size is $N \times C$.

\begin{figure}[t]
  \centering
  \includegraphics[trim={0 0 0 0mm},clip,width=1\linewidth]{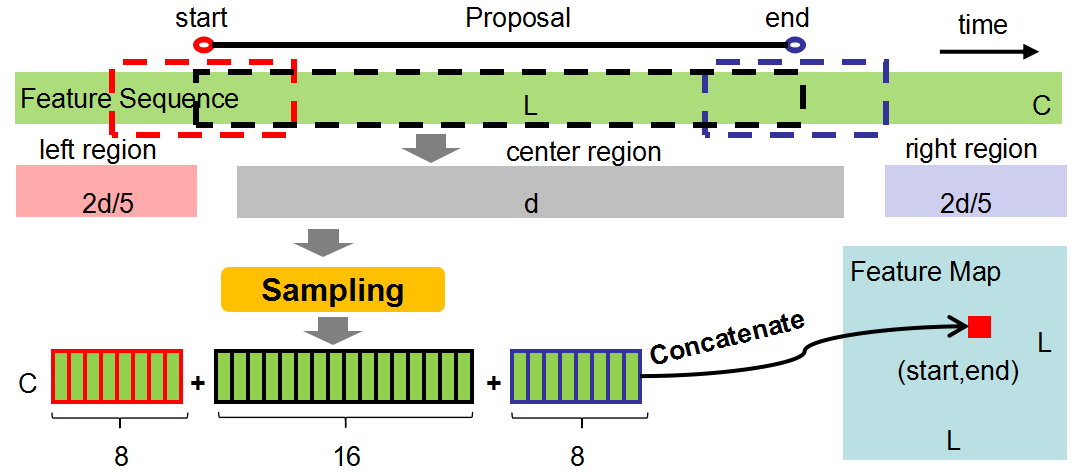}
  \caption{\small Details of the proposal feature generation layer. Given a feature sequence, we concatenate the sampled feature regions to construct proposal context feature map.
  }
\label{fig4}
\end{figure}

Fig. \ref{fig4} shows the detail of our PFG layer. First, for each candidate proposal $\varphi=(t_s, t_e)$, we sample $N_l$ locations from the left region $r^s=[t_s-d_g/k,t_s+d_g/k]$, $N_c$ locations from the center region $r^a=[t_s, t_e]$ and $N_r$ locations from the right region $r^e=[t_e-d_g/k,t_e+d_g/k]$ by linear interpolation, respectively, where $d_g=t_e-t_s, k=5$ and $N=N_l+N_c+N_r$. Then, with these sampling locations, we concatenate the corresponding temporal location features to produce the context proposal feature. Therefore, it is obvious to generate each proposal feature $f^p_{t_s,t_e}$ from the input feature $f^{in}$ through the following formula:
\begin{equation}
    f^p_{t_s, t_e, n, c} = w_l f^{in}_{t_l,c} + w_r f^{in}_{t_r,c},
\label{prop_feat}
\end{equation}
where
\begin{equation}
    t_l=\left\{
    \begin{array}{lr}
        \lfloor t_s-\frac{d_g}{k} + \frac{2d_g}{k(N_l-1)}n \rfloor, & n < N_l, \\
        \lfloor t_s + \frac{d_g}{N_c-1}n \rfloor, & N_l \leq n < N_l+N_c, \\
        \lfloor t_e - \frac{d_g}{k} + \frac{2d_g}{k(N_r-1)}n \rfloor, & n \geq N_l+N_c,
    \end{array}
    \right.
\end{equation}
\begin{equation}
    w_l=\left\{
    \begin{array}{lr}
        t_r - t_s + \frac{d_g}{k}-\frac{2d_g}{k(N_l-1)}n, & n < N_l, \\
        t_r - t_s - \frac{d_g}{N_c-1}n, & N_l \leq n < N_l+N_c, \\
        t_r - t_e + \frac{d_g}{k}-\frac{2d_g}{k(N_r-1)}n, & n \geq N_l+N_c,
    \end{array}
    \right.
\end{equation}
\begin{equation}
t_r = 1+t_l , w_r=1-w_l.
\end{equation}

When calculating gradient for training PFG layer, $f^p_{t_s,t_e}$ is differentiable for $f^{in}$, and its differential formulas are:
\begin{equation}
    \frac{\partial f^p_{t_s,t_e,n,c}}{\partial f^{in}_{t_l,c}}=w_l, \frac{\partial f^p_{t_s,t_e,n,c}}{\partial f^{in}_{t_r,c}}=w_r.
\end{equation}

In our experiments, we set $N_l = N_r = 8$ and $N_c=16$, thus $N=32$. Note that if $t_s \geq t_e$, then the proposal feature $f^p_{t_s,t_e}$ will be zero.
\\
\textbf{Action-aware completeness regression.} The ACR branch receives actionness score feature as input and outputs action completeness map $P^c$ to estimate the overlap between candidate proposals and ground truth action instances. In ACR, we employ the PFG layer and several two-dimensional convolutional layers for each proposal to explore semantic information in the global proposal level. As show in Tab. \ref{table1}, the PFG layer can transfer temporal actionness score features $asf$ to three-dimensional proposal feature tensors, which are fed into multi two-dimensional convolutional layers to generate $L$$\times$$L$ action completeness maps, denoted as $P^c=F_{(Conv11, Conv12, Conv13)}(F_{PFG}(asf))$. For each location or proposal in the action completeness map, we use a smooth L1 regression loss to supervise $P^{c}$ to generate reliable action completeness score.\\
\textbf{Temporal boundary classification.} The TBC branch receives dual stream feature as input and outputs boundary confidence map $P^{s,e}$ to estimate the starting and ending probabilities for dense candidate proposals. Similar with ACR, TBC includes the PFG layer, a three-dimensional convolutional layer and several two-dimensional convolutional layers. As show in Tab. \ref{table1}, dual stream features $dsf$ from DSB is transfered by the PFG layer to four-dimensional proposal tensors. Multi convolutional layers are stacked to generate $L$$\times$$L$$\times$$2$ boundary confidence maps written by $P^{s,e}=F_{(Conv21, Conv22, Conv23)}(F_{PFG}(dsf))$. For each location or proposal in the boundary confidence map, we use the binary classification loss to supervise $P^{s,e}$ to predict precise temporal boundaries.

\section{Training and Inference}
To jointly learn action completeness map and boundary confidence map, a unified multi-task loss is further proposed. In inference, with three score maps generated by DBG, a score fusion strategy and Soft-NMS can generate dense proposals with confidence.

\subsection{Label and Loss}
Given the annotation $\psi_{g}=\{\varphi_{i}=(ts_{i},te_{i})\}^{N_{g}}_{i=1}$ of a video $V$, we compose actionness label $g^a$ for auxiliary DSB actionness classification loss, boundary label $g^s, g^e$ for TBC boundary classification loss, and action completeness label $g^c$ for ACR completeness regression loss. For a given ground truth action instance $\varphi=(t_s,t_e)$, we define its action region as $r^a_g=[t_s,t_e]$, starting region as $r^s_g=[t_s-d_t,t_s+d_t]$ and ending region as $r^e_g=[t_e-d_t,t_e+d_t]$, where $d_t$ is the two temporal locations intervals.\\
\textbf{DSB actionness classification.} For each temporal location $i$ within actionness score feature sequence $P^{a}$, we denote its region as $r_{i}=[i-d_t/2,i+d_t/2]$. Then, we calculate maximum overlap ratio IoR for $r_{i}$ with $r^a_g$, where IoR is defined as the overlap ratio with ground truth proportional to the duration of this region. If this ratio is bigger than an overlap threshold 0.5, we set the actionness label as $g^a_{i}=1$, else we have $g^a_{i}=0$. With three actionness probability sequences $P^{a}$, we can construct DSB actionness classification loss using binary logistic regression: \\
\begin{equation}\
  \mathcal{L}^a_{DSB} = \frac{1}{3L}\sum_{j=1}^{3}\sum_{i=1}^{L}g^a_i log(p^{a_j}_{i})+(1-g^{a}_i)log(1-p^{a_j}_{i}).
\end{equation} \\
\textbf{TBC boundary classification.} For each location $(i,j)$ within starting confidence map $P^{s}$ or ending confidence map $P^{e}$, we denote its starting region as $r^s_{i,j}=[i-d_t/2,i+d_t/2]$ and its ending region as $r^e_{i,j}=[j-d_t/2,j+d_t/2]$. Similar with above actionness label, we calculate the starting label $g^s_{i,j}$ for $r^s_{i,j}$ with $r^s_g$ and the ending label $g^e_{i,j}$ for $r^e_{i,j}$ with $r^e_g$. We also adopt binary logistic regression to construct the classification loss function of TBC for the starting and ending separately:
\begin{equation}\
\small
  \mathcal{L}^s_{TBC} = \frac{1}{L^2}\sum_{i=1}^{L}\sum_{j=1}^{L} g^s_{i,j} log(p^s_{i,j})+(1-g^s_{i,j})log(1-p^s_{i,j}),
\end{equation} \\
\begin{equation}\
\small
  \mathcal{L}^e_{TBC} = \frac{1}{L^2}\sum_{i=1}^{L}\sum_{j=1}^{L}g^e_{i,j} log(p^e_{i,j})+(1-g^e_{i,j})log(1-p^e_{i,j}).
\end{equation} \\
\textbf{ACR completeness regression.} For each location or proposal $(i,j)$ within action completeness map $P^{c}$ , we denote its region as $r_{i,j}=[i,j]$. For $r_{i,j}$, We caculate the maximum Intersection-over-Union (IoU) with all $r^{c}_{g}$ to generate completeness label $g^{c}_{i,j}$. With the action completeness map $P^{c}$ from ACR, we simply adopt smooth L1 loss to construct the ACR loss function:
\begin{equation}\
\small
  \mathcal{L}^c_{ACR} = \frac{1}{L^2}\sum_{i=1}^{L}\sum_{j=1}^{L}smooth_{L1}(p^c_{i,j}-g^c_{i,j}).
\end{equation}

Following BSN, we balance the effect of positive and negative samples for the above two classification losses during training. For regression loss, we randomly sample the proposals to ensure the ratio of proposals in different IoU intervals [0,0.2],[0.2,0.6] and [0.6,1] that satisfies 2:1:1. We use the above three-task loss function to define the training objective of our DGB as:
\begin{equation}\
\small
     \mathcal{L}_{DBG} =\lambda\mathcal{L}^a_{DSB}+\mathcal{L}^s_{TBC}+\mathcal{L}^e_{TBC}+ \mathcal{L}^c_{ACR},
\end{equation}
where weight term $\lambda$ is set to 2 to effectively facilitate the actionness score features.

\subsection{Prediction and Post-processing}
In inference, different from BSN, three actionness probability sequences from DSB will not participate in computation of the final proposal results. Based on three score maps from ACR and TBC, we adopt post-processing to generate dense proposals with confidence.\\
\textbf{Score map fusion.} To make boundaries smooth and robustness, we average boundary probability of these proposals sharing the same starting or ending location. For starting and ending score map $P^s$, $P^{e}$ from TBC, we compute each location or proposal boundary probability $P^s_{i,j}$ and $P^e_{i,j} $ by:
\begin{equation}\
\small
P^s_{i,j}= \frac{1}{L}\sum_{k=1}^{L}P^s_{i,k},
P^e_{i,j}= \frac{1}{L}\sum_{k=1}^{L}P^e_{k,j}.
\end{equation}

For each proposal $(i,j)$ whose starting and ending locations are $i$ and $j$, we fuse boundary probability with completeness score map $P^{c}$ to generate the final confidence score $P_{i,j}$:
\begin{equation}\
\small
   P_{i,j}= P^c_{i,j} \times P^s_{i,j} \times P^e_{i,j}.
\end{equation}

For the fact that the starting location is in front of the ending location, we consider the upper right part of the score map, and then get the dense candidate proposals set as $\psi_{p}=\{\varphi_{i}=(i,j,P_{i,j})\}^{i<=j<=L}_{i=1,j=1}$.\\
\textbf{Proposal retrieving.} The above proposal generation will produce dense and redundant proposals around ground truth action instances. Subsequently, we need to suppress redundant proposals by Soft-NMS, which is a non-maximum suppression by a score decaying function. After Soft-NMS step, we employ a confidence threshold to get the final sparse candidate proposals set as $\psi_{p}=\{\varphi_{i}=(s_i,e_i,P_i)\}^{N}_{i=1} $, where $N$ is the number of retrieved proposals.

\begin{table*}[t]
\centering
\small
\caption{Comparison between our approach and other state-of-the-art temporal action generation approaches on validation set and test set of ActivityNet-1.3 dataset in terms of AR@AN and AUC.}
\begin{tabular}{ccccccccc}
\toprule
Method       & TCN   & MSRA  & Prop-SSAD & CTAP  & BSN   & MGG   & BMN   & Ours  \\ \hline
AR@100 (val) & -     & -     & 73.01     & 73.17 & 74.16 & 74.54 & 75.01 & \textbf{76.65} \\
AUC (val)    & 59.58 & 63.12 & 64.40     & 65.72 & 66.17 & 66.43 & 67.10 & \textbf{68.23} \\
AUC (test)   & 61.56 & 64.18 & 64.80     & -     & 66.26 & 66.47 & 67.19 & \textbf{68.57} \\ \bottomrule
\end{tabular}
\label{table2}
\end{table*}

\section{Experiments}

\subsection{Evaluation Datasets}

\textbf{ActivityNet-1.3.} It is a large-scale dataset containing 19,994 videos with 200 activity classes for action recognition, temporal proposal generation and detection. The quantity ratio of training, validation and testing sets satisfies 2:1:1. \\
\textbf{THUMOS14.} This dataset has 1,010 validation videos and 1,574 testing videos with 20 classes. For the action proposal or detection task, there are 200 validation videos and 212 testing videos labeled with temporal annotations. We train our model on the validation set and evaluate on the test set.

\subsection{Implementation Details}
For video representation, we adopt the same two-stream network \cite{ANetChallengecorr2016} pretrained on ActivityNet-1.3 and parameter setting by following ~\cite{BMNarxiv2019,BSNeccv2018} to encode video features. For ActivityNet-1.3, we resize video feature sequence by linear interpolation and set $L=100$. For THUMOS14,  we slide the window on video feature sequence with $overlap=0.5$ and $L=128$. When training DBG, we use Adam for optimization. The batch size is set to 16. The learning rate is set to $10^{-3}$ for the first 10 epochs, and we decay it to $10^{-4}$ for another 2 epochs. For Soft-NMS, we set the threshold θ to 0.8 on the ActivityNet-1.3 and 0.65 on the THUMOS14. $\epsilon$ in Gaussian function is set to 0.75 on both temporal proposal generation datasets.

\subsection{Temporal Proposal Generation}
To evaluate the proposal quality, we adopt different IoU thresholds to calculate the average recall (AR) with average number of proposals (AN). A set of IoU thresholds [0.5:0.05:0.95] is used on ActivityNet-1.3, while a set of IoU thresholds [0.5:0.05:1.0] is used on THUMOS14. For ActivityNet-1.3, area under the AR vs. AN curve (AUC) is also used as the evaluation metrics.\\
\textbf{Comparison experiments.} We further compare our DBG with other methods on the validation set of ActivityNet-1.3. Tab. \ref{table2} lists a set of proposal genearation methods including TCN ~\cite{TCNiccv2017}, MSRA ~\cite{MSRAcvpr2017}, Prop-SSAD ~\cite{SSADacmmm2017}, CTAP ~\cite{CTAPeccv2018}, BSN ~\cite{BSNeccv2018}, MGG ~\cite{MGGcvpr19} and BMN ~\cite{BMNarxiv2019}. Our method achieves state-of-the-art performance and improves AUC from 67.10\% to 68.23\%, which demonstrates that our DBG can achieve an overall performance promotion of action proposal generation. Especially, with multiple video representation networks and multi-scale video features, our ensemble DBG achieves $\bf 73.05$\% AUC, which ranks top-$1$ on ActivityNet Challenge $2019$ on temporal action proposals.

\begin{table}[t]
\caption{Comparison between DBG with other state-of-the-art methods on THUMOS14 in terms of AR@AN. }
\small
\centering
\setlength{\tabcolsep}{0.9mm}{
\begin{tabular}{clccccc}
\toprule
Feature & Method    & @50            & @100           & @200           & @500           & @1000          \\ \hline
C3D     & SCNN-prop & 17.22          & 26.17          & 37.01          & 51.57          & 58.20          \\
C3D     & SST       & 19.90          & 28.36          & 37.90          & 51.58          & 60.27          \\
C3D     & TURN      & 19.63          & 27.96          & 38.34          & 53.52          & 60.75          \\
C3D     & MGG       & 29.11          & 36.31          & 44.32          & 54.95          & 60.98          \\
C3D     & BSN+NMS   & 27.19          & 35.38          & 43.61          & 53.77          & 59.50          \\
C3D     & BSN+SNMS  & 29.58          & 37.38          & 45.55          & 54.67          & 59.48          \\
C3D     & BMN+NMS   & 29.04          & 37.72          & 46.79          & 56.07          & 60.96          \\
C3D     & BMN+SNMS  & \textbf{32.73} & 40.68          & 47.86          & 56.42          & 60.44          \\ \hline
C3D     & Ours+NMS  & 32.55          & \textbf{41.07} & \textbf{48.83} & \textbf{57.58} & 59.55          \\
C3D     & Ours+SNMS & 30.55          & 38.82          & 46.59          & 56.42          & \textbf{62.17} \\ \hline \hline
2Stream & TAG      & 18.55          & 29.00          & 39.61          & -              & -              \\
Flow    & TURN      & 21.86          & 31.89          & 43.02          & 57.63          & 64.17          \\
2Stream & CTAP      & 32.49          & 42.61          & 51.97          & -              & -              \\
2Stream & MGG       & 39.93          & 47.75          & 54.65          & 61.36          & 64.06          \\
2Stream & BSN+NMS   & 35.41          & 43.55          & 52.23          & 61.35          & 65.10          \\
2Stream & BSN+SNMS  & 37.46          & 46.06          & 53.21          & 60.64          & 64.52          \\
2Stream & BMN+NMS   & 37.15          & 46.75          & 54.84          & 62.19          & 65.22          \\
2Stream & BMN+SNMS  & 39.36          & 47.72          & 54.70          & 62.07          & 65.49          \\ \hline
2Stream & Ours+NMS  & \textbf{40.89} & \textbf{49.24} & \textbf{55.76} & 61.43          & 61.95          \\
2Stream & Ours+SNMS & 37.32          & 46.67          & 54.50          & \textbf{62.21} & \textbf{66.40} \\ \bottomrule
\end{tabular}}
\label{table3}
\end{table}

\begin{table}[t]
\centering
\small
\caption{Efficiency comparison among DBG and BMN and BSN in validation set of ActivityNet-1.3. \textit{e2e} means the method is able to be trained end-to-end. }
\begin{tabular}{cccccc}
\toprule
Method & \textit{e2e} & AR@100         & AUC            & $T_{pro}$      & $T_{all}$      \\ \hline
BSN    & $\times$            & 74.16          & 66.17          & 0.624          & 0.629          \\
BMN    & \checkmark            & 75.01          & 67.10          & 0.047          & 0.052          \\
DBG    & \checkmark            & \textbf{76.65} & \textbf{68.23} & \textbf{0.008} & \textbf{0.013} \\ \bottomrule
\end{tabular}
\label{efficiency table}
\end{table}

\begin{figure}[t]
  \centering
  \includegraphics[trim={0 0 0 0mm},clip,width=1\linewidth]{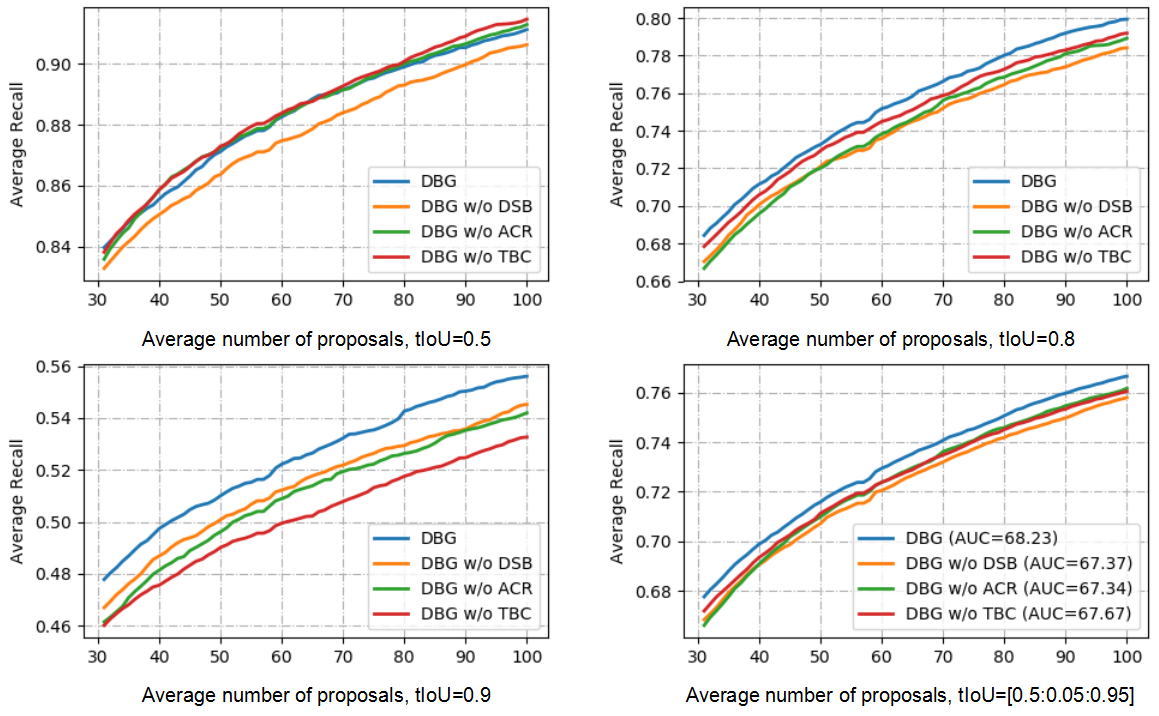}
  \caption{\small Ablation study of effectiveness of modules in DBG on validation set of ActivityNet-1.3 in terms of AR@AN curve.
  }
\label{fig5}
\end{figure}

\begin{table}[t]
\centering
\small
\caption{Performance analysis of PFG layer.}
\setlength{\tabcolsep}{1.4mm}{
\begin{tabular}{lcccccc}
\toprule
$N_l$/$N_c$/$N_r$ & 4/8/4 & 6/12/6         & 8/16/8         & 10/20/10 & 0/16/0 & 8/0/8 \\ \hline
AR@10             & 57.22 & \textbf{57.29} & \textbf{57.29} & 57.09    & 55.74  & 56.85 \\
AR@50             & 71.13 & 71.57          & \textbf{71.59} & 71.36    & 70.29  & 71.17 \\
AR@100            & 76.14 & 76.27          & \textbf{76.65} & 76.50    & 75.53  & 76.13 \\
AUC               & 67.91 & 68.14          & \textbf{68.23} & 68.11    & 66.94  & 67.83 \\ \bottomrule
\end{tabular}}
\label{pfg table}
\end{table}

Tab. \ref{table3} compares proposal generation methods on the testing set of THUMOS14. To ensure a fair comparison, we adopt the same video feature and post-processing step. Tab. \ref{table3} shows that our method using C3D or two-stream video features outperforms other methods significantly when the proposal number is set within [50,100,200,500,1000].

We conduct a more detailed comparison on the validation set of ActivityNet-1.3 to evaluate the effectiveness and efficiency among BSN, BMN, and DBG. As shown in Tab. \ref{efficiency table}, for a 3-minute video processed on Nvidia GTX 1080Ti, our inference speed accelerates a lot. And our proposal feature generation is reduced from 47ms to 8ms, while the total inference time decreases to 13ms.\\
\textbf{Ablation study.} We further conduct detailed ablation study to evaluate different components of the proposed framework, including DSB, ACR, and TBC, include the following\\
\emph{DBG w/o DSB}: We discard DSB and feed concatenated spatial and temporal features into the BSN-like BaseNet. \\
\emph{DBG w/o ACR}: We discard action-aware feature and auxiliary actionness classification loss, and adopt dual stream feature for action-aware completeness regression like TBC. \\
\emph{DBG w/o TBC}: We discard the whole temporal boundary classification module, and instead predict boundary probability sequence like actionness feature sequence in DSB.

\begin{figure}[t]
  \centering
  \includegraphics[trim={0 0 0 0mm},clip,width=1\linewidth]{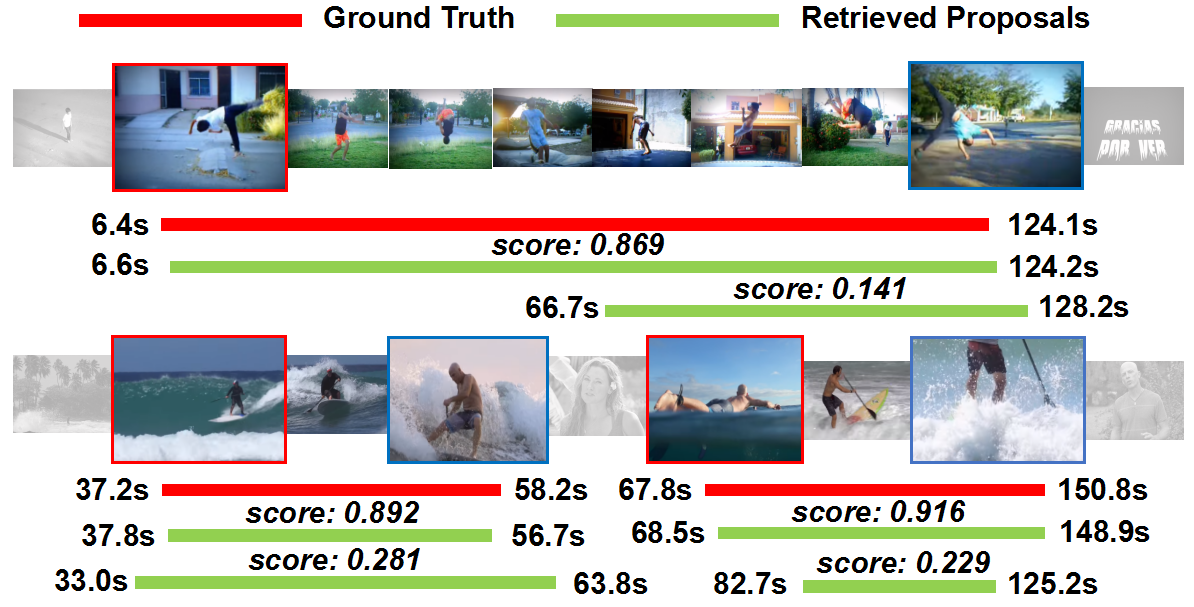}
  \caption{\small Visualization examples of proposals generated by DBG on ActivityNet-1.3 dataset.
  }
\label{fig6}
\end{figure}

As illustrated in Fig. \ref{fig5}, the proposed DBG outperforms all its variants in terms of AUC with different IoU thresholds, which verifies the effectiveness of our contributions. The \emph{DBG w/o ACR} results demonstrate that action-aware feature using auxiliary supervision is more helpful than dual stream feature for action completeness regression. The \emph{DBG w/o TBC} results explain the remarkable superiority of dense boundary maps for all proposals. When the IoU threshold is strict and set to be 0.9 for evaluation, a large AUC gap between \emph{DBG} (blue line) and \emph{DBG w/o TBC} (red line) shows TBC can predict more precise boundaries. Fig. \ref{fig6} shows more examples to demonstrate the effects of DBG on handling actions with various variations.\\
\textbf{Analysis of PFG layer.} To confirm the effect of the PFG layer, we conduct experiments to examine how different sampling locations within features affect proposal generation performance. As shown in Tab. \ref{pfg table}, The experiments that sampling 8, 16, 8 locations from left region, center region and right region respectively within proposal features achieves the best performance. The \emph{0/16/0} results indicate that context information around proposals are necessary for better performance on proposal generation. The \emph{8/0/8} experiment that only adopting left or right local region features for TBC to predict starting or ending boundary confidence map shows the importance of the global proposal information.\\
\textbf{Generalizability.} Following BMN, we choose two different action subsets on ActivityNet-1.3 for generalizability analysis: ``Sports, Exercise, and Recreation'' and ``Socializing, Relaxing, and Leisure'' as seen and unseen subsets, respectively. We employ I3D network ~\cite{I3Dcvpr2017} pretrained on Kinetics-400 for video representation. Tab. \ref{generalization table} shows the slight AUC drop when testing the unseen subset, which clearly explains that DBG works well to generate high-quality proposals for unseen actions.

\subsection{Temporal Proposal Detection}
To evaluate the proposal quality of DBG, we put proposals in a temporal action detection framework. We adopt mean Average Precision (\textit{m}AP) to evaluates the temporal action detection task. We adopt a set of IoU thresholds \{0.3,0.4,0.5,0.6,0.7\} for THUMOS14.

\begin{table}[t]
\centering
\small
\caption{Generalization evalation on ActivityNet-1.3.}
\begin{tabular}{lcccc}
\toprule
              & \multicolumn{2}{c}{Seen} & \multicolumn{2}{c}{Unseen} \\ \hline
Training Data & AR@100      & AUC        & AR@100       & AUC         \\ \hline
Seen+Unseen   & 73.30       & 66.57      & 67.23        & 64.59       \\
Seen          & 72.95       & 66.23      & \textbf{64.77}        & \textbf{62.18}       \\ \bottomrule
\end{tabular}
\label{generalization table}
\end{table}

\begin{table}[t]
\small
\centering
\caption{Action detection results on testing set of THUMOS14 in terms of \textit{m}AP@\textit{tIoU}.}
\begin{tabular}{lllllll}
\toprule
Method & classifier & 0.7           & 0.6           & 0.5           & 0.4           & 0.3           \\ \hline
SST    & SCNN-cls   & -             & -             & 23.0          & -             & -             \\
TURN   & SCNN-cls   & 7.7           & 14.6          & 25.6          & 33.2          & 44.1          \\
BSN    & SCNN-cls   & 15.0          & 22.4          & 29.4          & 36.6          & 43.1          \\
MGG    & SCNN-cls   & 15.8          & 23.6          & 29.9          & 37.8          & 44.9          \\
BMN    & SCNN-cls   & 17.0          & 24.5          & 32.2          & 40.2          & 45.7          \\
Ours   & SCNN-cls   & \textbf{18.4} & \textbf{25.3} & \textbf{32.9} & \textbf{40.4}     & \textbf{45.9}     \\ \hline
SST    & UNet       & 4.7           & 10.9          & 20.0          & 31.5          & 41.2          \\
TURN   & UNet       & 6.3           & 14.1          & 24.5          & 35.3          & 46.3          \\
BSN    & UNet       & 20.0          & 28.4          & 36.9          & 45.0          & 53.5          \\
MGG    & UNet       & 21.3          & 29.5          & 37.4          & 46.8          & 53.9          \\
BMN    & UNet       & 20.5          & 29.7          & 38.8          & 47.4          & 56.0          \\
Ours   & UNet       & \textbf{21.7} & \textbf{30.2} & \textbf{39.8} & \textbf{49.4} & \textbf{57.8} \\ \bottomrule
\end{tabular}
\label{thumos detection table}
\end{table}

We follow a two-stage ``detection by classifying proposals'' framework in evaluation, which feeds the detected proposals into the state-of-the-art action classifiers SCNN ~\cite{S-CNNcvpr2016} and UntrimmedNet ~\cite{UNetcvpr2017}. For fair comparisons, we use the same classifiers for other proposal generation methods, including SST ~\cite{SSTcvpr2017}, TURN ~\cite{TURNiccv2017}, CTAP ~\cite{CTAPeccv2018}, BSN ~\cite{BSNeccv2018}, MGG ~\cite{MGGcvpr19} and BMN ~\cite{BMNarxiv2019}.
The experimental results on THUMOS14 are shown in Tab. \ref{thumos detection table}, which demonstrates that DBG based detection significantly outperforms other state-of-the-art methods in temporal action detection methods. Especially, with the same IOU threshold 0.7, our DBG based detection achieves an \textit{m}AP improvements of 1.4\% and 1.2\% for two types of classifiers separately from BMN based methods.

\section{Conclusion}
This paper introduces a novel and unified temporal action proposal generator named Dense Boundary Generator (DBG). In this work, we propose dual stream BaseNet to generate two different level and more discriminative features. We then adopt a temporal boundary classification module to predict precise temporal boundaries, and an action-aware completeness regression module to provide reliable action completeness confidence. Comprehensive experiments are conducted on popular benchmarks including ActivityNet-1.3 and THUMOS14, which demonstrates the superiority of our proposed DBG compared to state-of-the-art methods.

\small\bibliographystyle{aaai}\bibliography{ref}

\end{document}